\pdfoutput=1

\documentclass[11pt]{article}

\usepackage[]{acl}

\usepackage{times}
\usepackage{latexsym}
\usepackage{graphicx}
\usepackage{amsmath}
\usepackage{url}
\usepackage{graphicx}
\usepackage{subfigure} 
\usepackage{multirow}
\usepackage{makecell}
\usepackage{colortbl}
\usepackage{amsfonts}

\usepackage{amssymb}
\usepackage{booktabs}
\usepackage{colortbl}
\usepackage{makecell}
\usepackage{verbatim}
\usepackage{latexsym}
\usepackage{graphicx,stackengine,scalerel}

\usepackage{stmaryrd}
\usepackage{helvet}
\usepackage{courier}
\usepackage{amsmath}
\usepackage{url}
\usepackage{multirow}
\usepackage{booktabs}
\usepackage{enumitem}
\usepackage{amssymb}
\usepackage{marvosym}
\usepackage{color}
\usepackage{caption}
\usepackage{titlesec}
\usepackage{floatrow}
\usepackage{caption}
\usepackage{rotating}
\usepackage{pifont}

\newfloatcommand{capbtabbox}{table}[][\FBwidth]

\usepackage[T1]{fontenc}

\usepackage[utf8]{inputenc}
\usepackage{verbatim}
\usepackage{enumitem}
\usepackage{array}

\usepackage{booktabs}
\usepackage{pgffor}

\usepackage{microtype}

\newcommand{\af}[1]{\textcolor{blue}{#1}}
\definecolor{darkgreen}{rgb}{0,0.35,0}
\newcommand{\xiyan}[1]{\textcolor{darkgreen}{#1}}

\definecolor{shallow_grey}{RGB}{167,194,203}
\definecolor{dark_grey}{RGB}{88,98,103}
\definecolor{light_green}{RGB}{66,195,183}
\definecolor{dark_blue}{RGB}{87,133,149}

\definecolor{dark_red}{RGB}{192,1,0}
\definecolor{dark_orange}{RGB}{255,147,2}
\definecolor{dark_yellow}{RGB}{255,213,121}

\definecolor{table_yellow}{RGB}{252,241,212}
\definecolor{table_green}{RGB}{198,233,230}
\definecolor{table_grey}{RGB}{197,200,202}

\newcommand\tablegreen[1]{\cellcolor{light_green!35}{#1}}
\newcommand\tableyellow[1]{\cellcolor{dark_yellow!35}{#1}}
\newcommand\tablegrey[1]{\cellcolor{dark_grey!35}{#1}}
\title{The Mystery of Compositional Generalization in Graph-based Generative Commonsense Reasoning}

\author{Xiyan Fu \\
  Dept. of Computational Linguistics \\
  Heidelberg University \\
  \texttt{fu@cl.uni-heidelberg.de} \\\And
  Anette Frank \\
  Dept. of Computational Linguistics \\
  Heidelberg University \\
  \texttt{frank@cl.uni-heidelberg.de} \\}

\begin{document}
\maketitle
\begin{abstract}
While LLMs have emerged as performant architectures for reasoning tasks, their compositional generalization capabilities have been questioned. 
In this work, we introduce a Compositional Generalization Challenge for Graph-based Commonsense Reasoning (CGGC) that goes beyond previous evaluations that are based on sequences or tree structures -- and instead involves a reasoning graph:
It requires models to generate a natural sentence based on given concepts and a corresponding reasoning graph,
where the presented graph involves a previously unseen combination of relation types. 
To master this challenge, models need to learn how to reason over relation tuples within the graph, and how to compose them when conceptualizing a verbalization.
We evaluate seven well-known LLMs using in-context learning and find that performant LLMs still struggle in compositional generalization. We investigate potential causes of this gap by analyzing
the structures of reasoning graphs, and find that different structures present varying levels of difficulty for compositional generalization. Arranging 
the order of demonstrations according to the 
structures' difficulty 
shows that organizing samples in an easy-to-hard schema 
enhances the compositional generalization ability of LLMs. \footnote{Code \& data: https://github.com/Heidelberg-NLP/CGGC}
\end{abstract}

\section{Introduction}

Reasoning \citep{brachman2004knowledge} has been widely explored and extended to a wide variety of situations involving logical or commonsense reasoning \citep{rashkin-etal-2018-modeling, talmor-etal-2019-commonsenseqa, bisk2020piqa}. Recently, LLMs such as GPT-3 \citep{brown2020language} and Llama2 \citep{touvron2023llama} have demonstrated astonishing performance on reasoning tasks \citep{lourie2021unicorn}. 

However, existing works found that LLMs are limited in 
scenarios that require generalization abilities, such as out-of-domain \citep{shen2021generalization, wang-etal-2021-language-models}, low-resource \citep{klein-nabi-2021-towards} and complex compositional \citep{dziri2024faith} tasks. 
\citet{hupkes2023taxonomy} concluded that inferior performance of models in such cases
originates from a lack of compositional generalization ability --
the ability to infer, from known components,  
a potentially inﬁnite number of novel combinations suitable to solve the given task.
With this ability, LLMs are 
expected to generalize to unseen and more complex reasoning scenarios without relying on large amounts of training instances.

To explore the compositional generalization abilities of LLMs in reasoning, existing works introduce benchmarks across various domains involving different data representations, such as natural language \citep{liu-etal-2022-challenges,yanaka-etal-2020-neural, fu-frank-2023-seti} and tree-based structures \citep{saparov2023testing, fu2023dynamic, kudo-etal-2023-deep}. These works facilitate the compositional generalization exploration in reasoning and have shown that LLMs are able to generalize to some extent, while being limited 
in
specific circumstances. However, to date, we note
a gap regarding the evaluation of compositional generalization abilities in the context of
graph-based reasoning. 
Graphs, as commonly used in real-world applications, offer flexible and diverse reasoning paths.
Recent evidence \citep{besta2024graph} suggests that graphs enhance LLM reasoning by enabling the use and combination of diverse reasoning paths.

\begin{figure*}
    \centering
    \includegraphics[width=\linewidth]{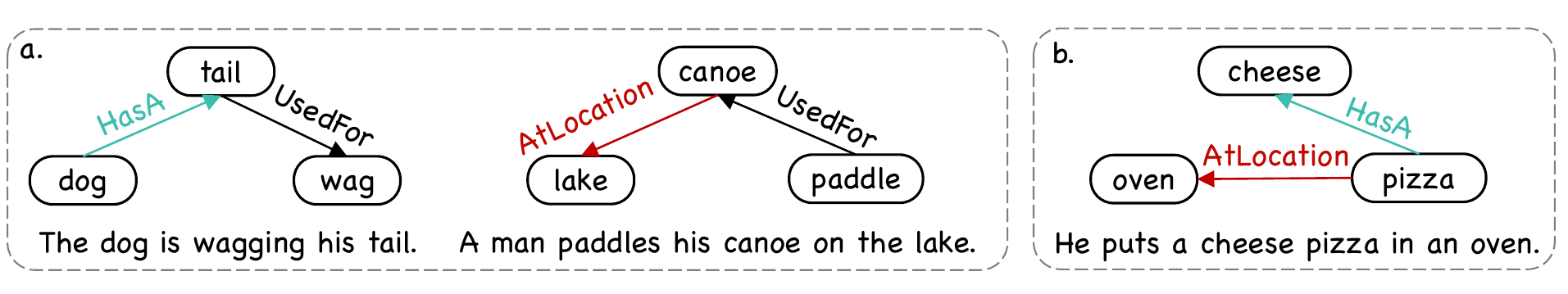}
    \caption{An instance of Compositional Generalization in Graph-based Commonsense Reasoning (CGGC). A model is expected to solve a test sample (b, \textit{composition}) that presents an input graph with an \textit{unseen} combination of relation types 
    (here: \textcolor{light_green}{HasA}\&\textcolor{dark_red}{AtLocation}). The ICL  demonstrations of the task in (a), by contrast, show each relation primitive
    in combination with other relation types, here:
    \textcolor{light_green}{HasA}\&UsedFor and 
    \textcolor{dark_red}{AtLocation}\&UsedFor.}
    \label{fig:intro}
\end{figure*}

Our work fills this gap by proposing 
CGGC, a \underline{C}ompositional \underline{G}eneralization Challenge for \underline{G}raph-based \underline{C}ommonsense Reasoning. CGGC builds on
the 
generative commonsense reasoning task 
CommonGen \cite{lin-etal-2020-commongen}, which tasked models to generate a coherent natural language sentence from a \textit{set of given concepts}. CGGC extends this task by requiring models to reason over a set of concepts presented in a connected graph structure. 
Fig.\ref{fig:intro}.b shows an example where a model is expected to generate a sentence such as `He puts a cheese pizza in an oven' from 
a knowledge graph that contains the set of target concepts \{cheese, oven, pizza\}.
For
this task, we design the compositional generalization test CGGC: The 
core idea of the CGGC challenge is to probe models 
on specific relationship compositions that have not previously been seen in the learning sets.
Fig.\ \ref{fig:intro} illustrates how 
an unseen combination of relations types must be jointly verbalized in a sentence, whereas each of the primitives has been seen in combination with other relation types.
\if false
be compositionally inferred from seen primitives in other combinations.
\fi
In the example, the model is required to reason over
an unseen relation combination, here, \textcolor{dark_red}{AtLocation}\&\textcolor{light_green}{HasA} -- while each of these primitive relations has been seen 
in combination with other relation types,
here:
\textcolor{dark_red}{AtLocation} in \textcolor{dark_red}{AtLocation}\&UsedFor, and \textcolor{light_green}{HasA} in \textcolor{light_green}{HasA}\&UsedFor. 

With the 
CGGC challenge, we systematically measure a model's compositional generalization ability in graph-based commonsense reasoning in an 
in-context learning (ICL) regime \citep{brown2020language}. Empirical results for seven well-known LLMs reveal challenges in compositionally generalizing to novel subgraph configurations. 
We analyze the factors that impact
compositional generalization in such cases by \textit{examining error trends} that change as a function of: 
i) \textit{compositions.} Focusing on the \textit{structure} of compositional reasoning graphs, we identify different  
schemas that result from composing 
primitive 
relations. Experiments 
show 
varying performance across different graph structures. E.g., relation compositions with uniform directionality
seem more straightforward compared to 
compositions that end in a common target node, or that start from a common node but end in distinct nodes (Fig.\ \ref{fig:subgraph}).
ii) \textit{primitives}. We analyze performances based on the distribution of primitives according to different relation types.
We find that LLMs more easily generalize to compositions involving 
common and frequent primitive relations. 

Given the observed performance variations, we 
arrange the order of presentation of graph structures in
ICL demonstrations 
according to their degree of difficulty.
Results indicate that ordering ICL demonstrations in an easy-first manner
enhances the models' compositional generalization ability.

\section{Related Work}

\paragraph{Analyzing 
Commonsense Reasoning} 
Existing analyses 
of commonsense reasoning 
focus on
representation \citep{zhou2020evaluating, su-etal-2022-mico}, interpretability \citep{rajani-etal-2019-explain}, bias \citep{sotnikova-etal-2021-analyzing, an-etal-2023-sodapop} and consistency \citep{maler2023evaluating}. Further, \citet{davison-etal-2019-commonsense, petroni-etal-2019-language, singh-etal-2023-viphy} probed language models for commonsense knowledge.
Others construct complex reasoning scenarios such as logical queries on commonsense knowledge graphs \citep{fang-etal-2024-complex} and geometric knowledge reasoning \citep{ding-etal-2024-knowledge}.
Commonsense reasoning has also been analyzed with downstream tasks, such as machine translation \citep{he-etal-2020-box, liu-etal-2023-revisiting-commonsense}, temporal question answering \citep{jain-etal-2023-language-models, wenzel2023overview}, etc.

Beyond these perspectives, generalization is another important
research direction. Existing works have shown that language models suffer from overfitting and are limited in generalization to out-of-domain examples \citep{sen-saffari-2020-models, kejriwal2020fine}, novel answers \citep{ma-etal-2021-exploring}, and various tasks \citep{zhang-etal-2023-cikqa}. In addition, 
\citet{shwartz-choi-2020-neural} found that LLMs tend to overestimate and amplify biases in training data. Our work extends the research and analyses of generalization in commonsense reasoning, 
from the perspective of compositional generalization.

\begin{table*}
    \centering
    \footnotesize 
    \resizebox{\columnwidth}{!}{
    \begin{tabular}{@{}llc@{}} \toprule
        Task \& Works& Examples  &Rep \\ \midrule
        \makecell[l]{\textbf{Question Answering} \\ \citet{liu-etal-2022-challenges}} &\makecell[l]{\underline{train}: Cow is a national animal of which country?  When did pandas come to USA? \\ \underline{test}: Panda is a national animal of which country?}  & \makecell[c]{natural \\ language}\\ \midrule
        \makecell[l]{\textbf{NLI} \\ \citet{yanaka-etal-2020-neural} \\ \citet{fu-frank-2023-seti}} & \makecell[l]{\underline{train}: He realizes a woman is smiling $\rightarrow$ A woman is smiling \\ A woman is smiling $\nrightarrow$ A man is smiling \\\underline{test}: He realizes a woman is smiling $\nrightarrow$ A man is smiling.} & \makecell[c]{natural \\ language} \\  \midrule
        \makecell[l]{\textbf{Deductive Reasoning} \\ \citet{fu2023dynamic} \\ \citet{saparov2024testing}} & \makecell[l]{\underline{train}: Alex is a dog. All dogs are mammals. $\rightarrow$ Alex is a mammal. \\ Fae is a cat. Fae is soft. $\rightarrow$ Fae is soft and a cat. \\ \underline{test}: Alex is a dog.  All dogs are mammals. Alex is not mean. $\rightarrow$ Alex is a mammal and not mean.} & \makecell[c]{tree} \\ \midrule
        \makecell[l]{\textbf{Commonsense} \\ \textbf{Reasoning} \\ (ours)} & \makecell[l]{\underline{train}: (dog, tail, HasA), (tail, wag, UsedFor) $\rightarrow$ The dog is wagging his tail. \\ (paddle, lake, AtLoc), (paddle, canoe, UsedFor) $\rightarrow$ A man paddles his canoe on the lake. \\ \underline{test}: (cheese, pizza, AtLoc), (pizza, cheese, HasA) $\rightarrow$ He puts a cheese pizza in an oven.} & \makecell[c]{graph} \\ 
\bottomrule
    \end{tabular}}
    \caption{Comparison of tasks exploring compositionality in reasoning. `Rep' shows the format of compositions.}
    \label{tab:composition_comparison}
\end{table*}

\paragraph{Compositional Generalization} Despite the success of LLMs on downstream tasks, their compositional generalization 
abilities are poorly understood
\citep{fodor1988connectionism, lake2017building, hupkes2020compositionality}.
Prior works have evaluated aspects of compositionality in PLMs
in semantic parsing \citep{lake2018generalization,kim-linzen-2020-cogs,qiu-etal-2022-evaluating}, machine translation \citep{li-etal-2021-compositional, dankers-etal-2022-paradox}, image caption generation \citep{nikolaus-etal-2019-compositional, bogin-etal-2021-covr}, etc., concluding
that state-of-the-art PLMs are still not able to perform compositional generalization.
To solve the issue, many approaches have been proposed, including data augmentation \citep{qiu-etal-2022-improving, levy-etal-2023-diverse}, specialized architectures \citep{zheng-lapata-2021-compositional-generalization, herzig-berant-2021-span,fu2023dynamic}, meta-learning \citep{conklin-etal-2021-meta, lake2023human}, etc. 

Recently, the exploration of compositional generalization in \textit{reasoning} has attracted increasing attention.
Existing works measure the 
compositional generalization abilities of models on
reasoning tasks such as question answering \citep{liu-etal-2022-challenges}, deductive reasoning \citep{saparov2023testing}, natural language inference \citep{yanaka-etal-2020-neural, fu-frank-2023-seti}, and arithmetic reasoning \citep{kudo-etal-2023-deep}. Our study differs from prior work in terms of representation types. 
We focus on the compositionality of reasoning on graph-based representations, which could facilitate
complex reasoning by offering 
diverse reasoning paths \citep{besta2024graph}. Table \ref{tab:composition_comparison} shows an overview of tasks with their underlying representations.
Our work is related to \citet{xu-etal-2023-compositional}, who propose to cluster predicates for compositional data-to-text generation, and who further test on compositions with \textit{more} predicates in novel domains. 
%
In contrast, we 
focus on \textit{novel} compositions by
\textit{recombining} known relations, serving as acomplementary work.

Unlike prior work 
we conduct detailed analyses of compositional graph structures, 
hence our findings can facilitate future 
reasoning tasks.

\paragraph{Generative Commonsense Reasoning} The CommonGen task proposed by \citet{lin-etal-2021-common}, aims to advance machine commonsense towards generative reasoning abilities. Based on this benchmark, prior works have improved generation quality by incorporating explicit knowledge \citep{liu2021kg, liu2022kgr4} and visualizing relational scenes \citep{wang2022contextualized}. More recent work focused on enhancing diversity \citep{liu-etal-2023-dimongen, zhang2024improving, jinnai2024generating} and robustness \citep{neerudu-etal-2023-robustness}. The task has also
been extended to other domains, such as testing negative knowledge \citep{chen-etal-2023-say} and visual commonsense generation \citep{tang-etal-2023-learning-imagine, cui2024more}. We complement these studies by providing a new perspective on the generalization ability of LLMs. 

CommonGen and CGGC are grounded in 
the same 
dataset, but differ in motivation and research fields: CommonGen tasks a model to generate a natural language sentence based on a set of given concepts. It advances machine commonsense toward generative reasoning abilities. We extended this dataset with reasoning graphs, and split it elaborately to test for the compositional  generalization abilities of models, in a generative graph-based commonsense reasoning task.  
Our benchmark CGGC offers the \textit{first graph-based compositional generalization evaluation benchmark}, 
enriching  the compositional generalization research field.

\section{Defining a new CGGC Challenge}
Our new challenge CGGC, aiming to test for Compositional Generalization abilities in  Graph-based generative Commonsense Reasoning tasks, extends the existing CommonGen 
task of \citet{lin-etal-2021-common}. 
CGGC asks systems to generate a plausible natural language sentence given \textit{a set of concepts} and \textit{a reasoning graph} that connects these concepts, showing how they can 
relate to each other. 
The 
sentence is expected to serve as a description covering commonsense relations between given concepts.

Given the novel task, we aim to explore whether and to what extent LLMs can perform compositional generalization -- which 
consists in
understanding and verbalizing novel \textit{compositions} of previously seen 
\textit{primitive relation types}, presented in a graph-structured representation. In our work we define a relation type in the reasoning graph as a primitive, and a compound that requires several relation types in the reasoning graph as a composition. Using \{\textit{oven, pizza, cheese}\} in Fig.\ \ref{fig:intro} as example, the relation compound \textcolor{dark_red}{AtLocation}\&\textcolor{light_green}{HasA} is regarded as a composition; its constituent relation types \textcolor{dark_red}{AtLocation} and \textcolor{light_green}{HasA} are seen as primitives.


\section{Datasets}
\label{sec:data_construction}
In this section we introduce the benchmark for our novel CGGC challenge. It relies on the CommonGen dataset \citep{lin-etal-2020-commongen} and the commonsense resource ConceptNet \citep{speer2017conceptnet} (Section \ref{sec:data_setup}). With these resources, we extend samples with reasoning graphs (Section \ref{sec:graph_construction}), and further split the constructed data based on the compositional generalization features for evaluation (Section \ref{sec:graph_label}).

\subsection{Dataset Pre-Processing}
\label{sec:data_setup}
CommonGen \citep{lin-etal-2020-commongen} \footnote{We use the train and dev data together, and ignore the test data given it is unavailable for reasons of leaderboard testing.} tasks models to generate a coherent sentence given a set of common concepts. For example, \{\textit{tail, dog, wag}\} $\rightarrow$ \textit{The dog is wagging his tail}. The input is an unordered set of $k$ concepts, denoted as $\mathcal{X}$ = \{$c_{1}$, $c_{2}$, ..., $c_{k}$\}. Each concept $c_{i}$ is a common object (noun) or action (verb), which is guaranteed to appear as a ConceptNet unigram \citep{speer2017conceptnet}. The expected output is a coherent sentence $\mathcal{Y}$ that describes a common scenario from daily life, using all given concepts in $\mathcal{X}$. 
Each reference consists of an average of 11 words and introduces about 3 new meaningful words not included in the given concept set (see the Appendix \ref{app:stat_target_sent} for details).

We enrich each concept set with a 
commonsense reasoning graph that provides related commonsense facts and relations.
As knowledge resource, we choose ConceptNet, as it encompasses all candidate concepts. ConceptNet nodes represent concepts, and its edges provide commonsense knowledge relations between them, covering
34 relation types. \footnote{\url{https://github.com/commonsense/conceptnet5/wiki/Relations}} To facilitate 
the following 
compositionality evaluations, we merge some relations with similar meanings (e.g., \{/r/InstanceOf, /r/MannerOf\} $\rightarrow$ /r/IsA) and ignore some infrequent relations (e.g., /r/LocatedNear, /r/SymbolOf). We finally select 13 frequent relations following \citep{becker-etal-2021-co}. More details are given in the Appendix \ref{app:cn_rels}.

\subsection{Graph Construction}
\label{sec:graph_construction}
To construct a ConceptNet-based 
reasoning graph 
that fits a given concept set and an associated target sentence $y$, following \citet{plenz-etal-2023-similarity}, we i) construct a similarity-weighted ConceptNet subgraph, that assigns weights to each triple based on their similarity to the reference sentence $y$,
and ii) apply Dijkstra’s algorithm \citep{dijkstra2022note} to find weighted shortest paths between all concept pairs.

\begin{figure}
    \centering
    \includegraphics[width=\linewidth]{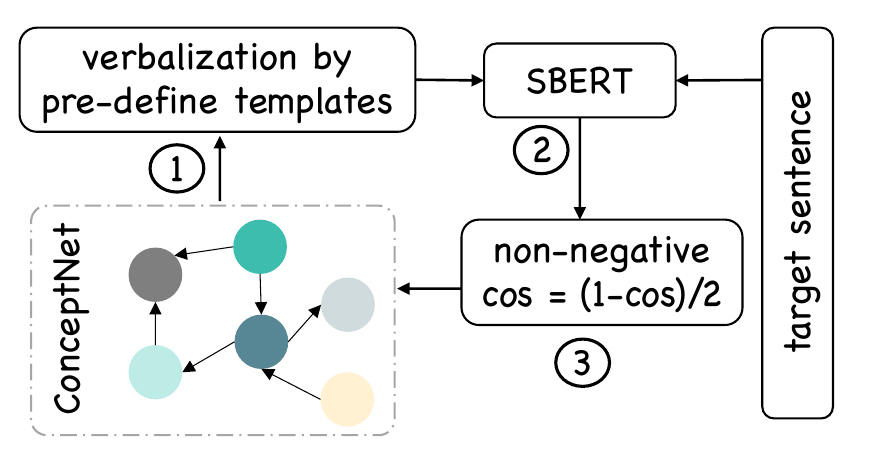}
    \caption{The process of constructing a reasoning graph given ConceptNet and a target sentence.}
    \label{fig:graph_constr}
\end{figure}

Fig. \ref{fig:graph_constr} illustrates this construction process: 
We verbalize the triplets in ConceptNet to sentences using pre-defined templates (\ding{172}) and further encode these sentences using S(entence-)BERT \citep{reimers-gurevych-2019-sentence} (\ding{173}). For instance, the triple (tail, UsedFor, wag) is verbalized as `Tail is used for wag'. We also 
encode a target sentence in $y$ using SBERT, to attain its representation. We calculate cosine similarity between all candidate triple representations and the target sentence representation. To provide non-negative weights for Dijkstra’s algorithm, we transform the calculated cosine similarity into a non-negative value, signifying semantic dissimilarity between concepts (\ding{174}). This conversion involves scaling the cosine similarity value to a new weight using the formula $\frac{1 - \text{cos}}{2}$. The resulting values serve as weights for all triples.

Hence, for each sample, we 
assign weights to triples based on the corresponding target sentences. The 
ensuing shortest subgraph calculation minimizes the cumulative edge weights, effectively maximizing semantic similarity across concepts and references. Ultimately, each sample will be assigned a 
reasoning graph \footnote{For a concept set with multiple reference sentences, we construct an individual reasoning graph 
for each reference.}. 
We conduct a human evaluation to assess the quality of the constructed graphs in Appendix \ref{app:graph_eval}.

\subsection{Compositional Generalization Label}
\label{sec:graph_label}
To perform compositional generalization evaluation on our dataset, we assign a graph label to each sample. 
The label indicates the types of relations that will be needed to infer and generate a sentence that entails the given concept set.
E.g., the graph label for the sample \{\textit{oven, cheese, pizza}\} shown in Fig. \ref{fig:intro} is \textcolor{dark_red}{AtLocation}\&\textcolor{light_green}{HasA}.

We define graph labels for all data samples and only select compositional labels which contain at least two distinct relations for compositional generalization tests. After filtering we select 468 compositional graph labels for our experiments. These labels are related to 26,819 examples in total, where each label corresponds to at least five samples. We categorize all selected data instances into three groups, based on the input concept set sizes. Table \ref{tab:statistics} provides statistics
of the data, with
further details about the 
extended 
reasoning graphs.\footnote{Note that during graph construction, intermediate concept nodes may be added. This can be seen in column 4 of Table \ref{tab:statistics}.}

\section{Experimental Setup}

\begin{table}
    \centering
    \resizebox{\columnwidth}{!}{
    \begin{tabular}{c|cccccc} \toprule
         \multirow{3}{*}{\thead[c]{Concepts \\ Sets }}&\multicolumn{3}{c}{general} &\multicolumn{3}{c}{graph} \\ \cmidrule(r){2-4} \cmidrule(r){5-7}
         & \#set & \#sent & \#len$_s$ & \#nodes & \#edges &\#rel\\ \midrule
        all &26819  &55269 &10.68 &9.07 &7.73 &4.63  \\ \midrule
        -size=3 &19684  &42747 &10.16 &8.31 &7.01 &4.41   \\
        -size=4 &3947  &8548 &11.75 &10.85 &9.43 &5.15   \\
        -size=5 &3188  &3972 &13.89 &13.33 &11.81 &5.76   \\ \bottomrule
    \end{tabular}}
    \caption{Data statistics for our novel CGGC benchmark. 
    We provide details about concept sets, reference sentences $s$, and the original vs.\ extended reasoning graphs.
   }
    \label{tab:statistics}
\end{table}


\subsection{Evaluated LLMs and Methods}

We select 7 open access autoregressive LLMs: i) Llama2, Llama2-chat \citep{touvron2023llama}; ii) Mistral-v0.1, Mistral-Instructv0.1 \citep{jiang2023mistral}; iii) Falcon, Falcon-instruct \citep{penedo2023refinedweb}; iv) GPT-J \citep{gpt-j}. The first three model types include both \textit{vanilla} pre-trained models and \textit{instruct} models fine-tuned on instruction datasets. 
We will refer to all instruction tuning-based model variants as x-chat, e.g., Llama2-chat. For all LLMs, we choose their 7 billion versions (GTP-J is aligned with 6B) to avoid the model scale effects. 

In addition, we evaluate GPT4 \citep{achiam2023gpt}, to also assess the compositional generalization abilities of a larger-scale LLMs.

In all experiments we use in-context learning with a fixed prompt as  task adaptation technique, since existing works have proved its superior effectiveness in compositional generalization \citep{qiu-etal-2022-evaluating, saparov2023testing}.\footnote{Details about prompts can be found in Appendix \ref{app:prompts}.} We give four demonstrations per prompt \footnote{The value is determined by the empirical experiments.} in our main experiments.

\begin{table}
    \centering
    \resizebox{\columnwidth}{!}{
    \begin{tabular}{@{}ccl@{~~}c@{}} \toprule
         &&sample& label \\ \toprule
         &\rotatebox{90}{\hspace*{-2mm}test}&\thead[l]{\{cheese, pizza, oven\} \\ $\rightarrow$ He puts a cheese pizza in an oven.} &{\textcolor{dark_red}{AtLocation}-\textcolor{light_green}{HasA}} \\  \toprule
         \multirow{6}{*}{\rotatebox{90}{demonstrations\hspace*{-3mm}}}&\hspace*{-2mm}\rotatebox{90}{id}&\thead[l]{\{cup, tea, table\} \\ $\rightarrow$ He puts a cup of tea on the table.}&\textcolor{dark_red}{AtLocation}-\textcolor{light_green}{HasA} \\ \cmidrule(r){2-4}
         &\multirow{4}{*}{\hspace*{-2mm}\rotatebox{90}{ood}} &\thead[l]{\{tail, dog, wag\} \\ $\rightarrow$ The dog is wagging his tail.} & UsedFor-\textcolor{light_green}{HasA}\\ \cmidrule(r){3-4}
         &&\thead[l]{\{lake, paddle, canoe\} \\ $\rightarrow$ A man paddles his canoe on the lake.} & \textcolor{dark_red}{AtLocation}-UsedFor  \\ \bottomrule
    \end{tabular}}
    \caption{Example of in-context learning evaluation. We show
    the test sample and corresponding in-distribution (id) and out-of-distribution (ood) demonstrations.}
    \label{tab:setting_example}
\end{table}

\subsection{Evaluation Data Split}
To conduct the compositional generalization evaluations, we control the dataset splits based on primitives and compositions. Specifically, we design two settings: i) \textbf{In-Distribution}. Demonstration samples and the evaluated samples come from the same distribution, meaning that 
all samples share the same reasoning graph label. ii) \textbf{Out-of-Distribution} (i.e., compositional generalization). Here, demonstration samples and evaluated samples come from different distributions. We guarantee that the primitive relation types in the evaluated sample are encountered in the demonstration samples. Table \ref{tab:setting_example} shows examples for both settings.

\begin{table*}
    \centering
    \resizebox{.7\columnwidth}{!}{
    \begin{tabular}{ll|ccccc|cc|c} \toprule
         Models & Dis &\multicolumn{2}{c}{ROUGE-2/L} &\multicolumn{2}{c}{BLEU-3/4} &METEOR &CIDEr &SPICE &Coverage \\ \midrule
         \multirow{3}{*}{Llama2} &id &8.57 &27.49 &14.00 &9.00 &16.19 &7.98 &24.30 &71.79 \\
          &ood &\tableyellow6.27 &\tableyellow24.57 &\tableyellow10.10 &\tableyellow5.80 &\tableyellow15.84 &\tableyellow6.48 &\tableyellow22.80 &\tableyellow67.75 \\  \cmidrule{2-10}
          &$\Delta$ &2.30 &\tablegrey2.92 &\tablegrey3.90 &\tablegrey3.20 &\tablegrey0.35 &\tablegrey1.50 &\tablegrey1.50 &\tablegrey 4.04 \\   \midrule
          \multirow{3}{*}{Llama2-chat} &id &8.35 &26.85 &13.60 &8.70 &16.13 &7.50 &24.00 &70.42 \\
          &ood &6.11 &23.03 &9.50 &5.30 &14.97 &5.09 &21.20 &65.64 \\   \cmidrule{2-10}
          &$\Delta$ &\tablegrey2.14 & 3.82 & 4.10 & 3.40 & 1.16 & 2.41 & 2.80 & 4.78 \\ \midrule
         \multirow{3}{*}{Mistral} &id &\tablegreen9.50 &\tablegreen28.79 &\tablegreen15.50 &\tablegreen10.40 &\tablegreen17.02 &\tablegreen8.60 &\tablegreen24.50 &69.92 \\	
          &ood &6.15 &23.69 &9.50 &5.50 &15.49 &5.85 &21.00 &59.88\\ \cmidrule{2-10}  
          &$\Delta$ &3.35 &5.10 &6.00 &4.90 &1.53 &2.75 &3.50 &10.04 \\ \midrule
         \multirow{3}{*}{Mistral-chat} &id &8.37 &26.86 &13.70 &8.70 &15.33 &7.67 &23.50 &\tablegreen71.89  \\
          &ood &5.56 &21.79 &9.10 &5.10 &13.45 &5.40 &19.10 &55.72 \\  \cmidrule{2-10} 
          &$\Delta$ &2.81 & 5.07 &4.60 &3.60 &1.88 &2.27 &4.40 & 16.17\\ \midrule
         \multirow{3}{*}{Falcon} &id &8.44 &26.97 &13.80 &9.00 &15.14 &7.59 &22.10 &64.06 \\ 
          &ood &5.99 &22.73 &9.80 &5.70 &13.51 &5.57 &19.10 &55.21\\ \cmidrule{2-10}  
          &$\Delta$ &2.45 &4.24 &4.00 &3.30 &1.63 &2.02 &3.00 &8.85 \\ \midrule
          \multirow{3}{*}{Falcon-chat} &id &7.42 &25.48 &12.30 &7.90 &14.31 &6.83 &20.10 &60.22 \\
          &ood &5.03 &21.24 &8.00 &4.50 &12.89 &4.95 &17.40 &50.55 \\  \cmidrule{2-10} 
          &$\Delta$ & 2.39 & 4.24 & 4.30 & 3.40 & 1.42 & 1.88 & 2.70 & 9.67\\ \midrule
          \multirow{3}{*}{GPT-J} &id &7.31 &24.40 &12.10 &7.60 &13.33 &6.37 &19.00 &52.20 \\
          &ood &5.10 &20.68 &8.00 &4.10 &12.56 &4.86 &16.60 &43.71 \\ \cmidrule{2-10} 
          &$\Delta$ & 2.21 & 3.72 & 4.10 & 3.50 & 0.77 & 1.51 & 2.40 & 8.49\\ \midrule\midrule
          \multirow{3}{*}{GPT-4o$^5$} &id &10.43 & 29.72&  15.70 &  10.70&17.78 & 9.75  & 37.70 & 97.68 \\
          &ood &8.46   & 27.04   &  12.40 &8.10  & 16.67 &  8.63 &35.50   & 95.17  \\ \cmidrule{2-10} 
          &$\Delta$ & 1.97 & 2.68 &  3.30 & 2.60 & 1.11& 1.12  & 2.20 &  2.51 \\ \bottomrule
    \end{tabular}}
    \caption{Performance of seven LLMs on the CGGC tasks in two configurations: \textit{in-distribution} (id) and \textit{compositional generalization} (ood). $\Delta$ indicates the gap between the setting of id and ood, calculated as $\Delta$ = id - ood. We highlight the maximum value of id and ood by \raisebox{0.2\baselineskip}{\colorbox{table_green}{}} and \raisebox{0.2\baselineskip}{\colorbox{table_yellow}{}} respectively, and the minimum value of $\Delta$ in \raisebox{0.2\baselineskip}{\colorbox{table_grey}{}}. Aggregated results are shown in the Appendix \ref{app:aggregated_rst} for a better illustration of comparisons among different models.}
    \label{tab:main_rst}
\end{table*}

\subsection{Evaluation Metrics}
\paragraph{Quality Evaluation} Following \citet{lin-etal-2020-commongen}, we use seven evaluation metrics from three categories, focusing on
i) \textit{surface similarity} by
concentrating on n-gram overlap between generations and references, using BLEU \citep{papineni-etal-2002-bleu}, ROUGE \citep{lin2004rouge} and METEOR \citep{banerjee-lavie-2005-meteor};
ii) \textit{concept associations}, 
assuming system generations and human references use similar concepts and focusing on evaluating the associations between mentioned concepts, using CIDEr \citep{vedantam2015cider} and SPICE \citep{anderson2016spice};
iii) \textit{task performance}, by
analyzing whether the model completes the given task. Here, Coverage \citep{lin2004rouge} calculates the average percentage of input concepts present in the lemmatized output. 

\paragraph{Human Evaluation} \label{huamn-eval}
Following \citep{lu-etal-2022-neurologic, meng2022controllable,zhang2023tractable}, we conduct a human evaluation of the generated sentences $y$ across four dimensions: i) \textit{Quality}: Is the sentence well-formed and fluent? ii) \textit{Plausibility}: Does the sentence describe a plausible situation? 
iii) \textit{Concepts}: Does the sentence include the given concepts in a meaningfull way? iv) \textit{Overall}: Considering the above three metrics, does the sentence meaningfully combine all given concepts into a well-formed scenario? For each aspect, annotators indicated their agreement with the pre-defined statement using the scale: \textit{Yes} (3 points), \textit{Somewhat} (2 points), and \textit{No} (1 point).
\footnote{Evaluation guidelines are provided in the Appendix \ref{app:huamn_eval_guidline}.}

\paragraph{Relation Verification} \label{verification} CGGC performs compositional generalization evaluation based on the reasoning relations, assuming that LLMs that understand them can use them 
for generating sentences. However, existing research indicates that LLMs might produce correct answers without applying the correct reasoning. In our context, this means a fitting sentence could be generated without utilizing the provided relations. To address this issue, we developed a \textit{verification model} to ensure that the generated sentences indeed incorporate the intended reasoning relations. 

For this purpose we
use Llama2 with a feed-forward classification layer to classify relation types based on two concepts and a target sentence. For example, given the concepts \{\textit{oven, pizza}\} and the sentence \textit{He puts a cheese pizza in an oven}, the model is expected to predict the relation \textit{/r/AtLocation}. To evaluate the model, we compare the predicted relation with those specified in the reasoning graph. If the prediction matches the provided relation, we consider it to be genuinely used, aligning with our expectations for compositional evaluation.\footnote{For details on this model, see Appendix \ref{app:verification}.} 
For further analyses, we use the \textit{verification model} to filter results, focusing solely on examples where all composed relations are accurately applied. After filtering, an average of 44.74\% and 44.81\% of the data is removed for in-distribution and compositional generalization, respectively.

\section{Results} \label{main_rst}
\subsection{Overall Results}
Table \ref{tab:main_rst} shows the performance of all LLMs in two data configurations: \textit{in-distribution} (id) and \textit{compositional generalization} (ood).\footnote{Due to the high cost of large-scale LLMs, we randomly sample 100 instances for GPT-4o's evaluation. For fair comparison, results of GPT-4o will not be compared for the following qualitative analysis.} We provide aggregated results in Appendix \ref{app:aggregated_rst} for better overview.

According to eight evaluation metrics, we observe that \textit{Mistral} and \textit{Llama2} generally achieve the best performance under in-distribution and compositional generalization settings, respectively (highlighted in green and yellow in Table \ref{tab:main_rst}). 
Across all evaluated models, including the powerful GPT-4, the gap ($\Delta$) between the two data configurations consistently shows positive values.
This suggests that the \textbf{evaluated LLMs still lack compositional generalization ability to some extent} when encountering unseen composition instances.
In addition, we find that various model groups show variance in absolute performance and the compositional generalization abilities.
Llama2 models show high absolute performance in both data configurations and superior compositionality. Mistral models achieve relatively high absolute performance but low compositionality, whereas GPT-J shows the opposite trend (Fig. \ref{app:fig_aggregrated_rst} in the Appendix \ref{app:aggregated_rst} shows details).

We also compare the performances of the vanilla 
vs.\ chat version for each model type,\footnote{GPT-J does not have a chat version, so we ignore it here.
} e.g., Llama2 vs. Llama2-chat. Results do not indicate a consistent trend:
For Llama2, the vanilla version shows superior compositionality in 7 metrics but inferior results in the remaining (ROUGE-2) metric, i.e., 7 vs.\ 1. This trend is not observed with Mistral and Falcon, where the metrics of superior and inferior results between vanilla and chat versions are more balanced: 3 vs.\ 5 for Mistral and 3 vs.\ 4 for Falcon.


\begin{table}
    \centering
    \resizebox{\columnwidth}{!}{
    \begin{tabular}{ll|cccc} \toprule
        Models & Dis& Quality & Plausibility & Concepts & Overall \\ \midrule 
        \multirow{2}{*}{Llama2} &id &2.39&2.43&2.29&2.22  \\ 
        &ood&1.96&1.61&2.02&1.95  \\ \midrule
        \multirow{2}{*}{Llama2-chat} & id&2.18 &2.41 &2.32 & 2.14 \\
        &ood &1.88 &1.93&1.87 & 1.99 \\ \midrule
        \multirow{2}{*}{Mistral} &id &2.20&2.20&2.28&2.10 \\
        &ood &2.02 &2.00 &1.90&1.80 \\ \midrule
        \multirow{2}{*}{GPT-4o} &id &2.54 &2.56& 2.94& 2.62\\
        &ood &2.26 &2.24 & 2.80 &2.46 \\
        \bottomrule
    \end{tabular}}
    \caption{Human evaluation results of four models on the CGGC task in two configurations: in-distribution (id) and compositional generalization (ood).}
    \label{tab:human}
\end{table}

\subsection{Human Evaluation}
To avoid the limitation of rigid automatic evaluation metrics, we also conduct human evaluations following \citep{meng2022controllable,zhang2023tractable}. We selected 50 samples from each of four representative models (maximum performance / minimum compositionality gap) under both settings (in-distribution and out-of-distribution), resulting in a total of 400 samples. These samples were mixed and presented to two annotators.

Table \ref{tab:human} shows human evaluation results. Comparing the results between the 
in-distribution (id) and compositional generalization (ood) settings, the four evaluated models represent higher values in id compared to ood setting. This trend is consistent with the automated metrics, reinforcing that LLMs still face challenges with compositional generalization. Notably, GPT-4 achieves the best performance across both configurations, outperforming other baselines by 0.4 and 0.47 points overall.

\begin{table*}
    \centering
    \resizebox{.85\columnwidth}{!}{
    \begin{tabular}{c|ccccc|cc|c|c} \toprule
         Type &\multicolumn{2}{c}{$\Delta$ROUGE-2/L($\downarrow$)} &\multicolumn{2}{c}{$\Delta$BLEU-3/4($\downarrow$)} &$\Delta$METEOR($\downarrow$) &$\Delta$CIDEr($\downarrow$) &$\Delta$SPICE($\downarrow$) &$\Delta$Coverage($\downarrow$) & Rank\\ \midrule
         A &2.28 &2.91 &2.63 &1.44 &0.07 &1.44 &0.78 &2.96 &\ding{172} \\
         B &2.58 &2.94 &3.07 &1.61 &0.09 &1.81 &1.00 &3.06 &\ding{173} \\ 
         C &2.72 &3.50 &3.09 &2.09 &0.11 &1.97 &1.51 &4.35 &\ding{174} \\ \bottomrule
    \end{tabular}}
    \caption{The performance \textbf{gap} between in-distribution and compositional generalization ($\Delta = \text{id} - \text{ood}$) for three reasoning graph structures (shown in Fig. \ref{fig:subgraph}). \textit{Rank} indicates the difficulty levels, calculated by the performance gap.}
    \label{tab:rst_type}
\end{table*}

\section{Error Analysis}
\label{analyses}
In this section we investigate potential causes of limitations in compositional generalization, by analyzing error trends in relation to compositions (Section \ref{ana:compos}) and primitives (Section \ref{ana:prim}). 
\footnote{Results of Llama2 are used for further analysis, given its superior performance. For each analysis, we perform three runs with different seeds.} 

\subsection{Composition Analysis}
\label{ana:compos}

\begin{figure}
    \centering
    \includegraphics[width=.9\linewidth]{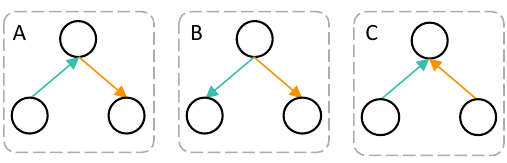}
    \caption{Different types of compositional connection schemas.  \raisebox{0.2\baselineskip}{\colorbox{light_green}{}} and \raisebox{0.2\baselineskip}{\colorbox{dark_orange}{}} arrows ($\rightarrow$) denote primitive relations. 
    }
    \label{fig:subgraph}
\end{figure}

\paragraph{Graph Structure} We examine the \textit{graph structures} that indicate 
how the composition is structured.
Considering the complexity of compositions with multiple primitives, we constrain the experiment to two primitives. \footnote{Experimental data is sampled from graphs of original concept set size 3. For more details see Appendix \ref{app:stat_graph_type}.} Graphs composed of two primitives (i.e., connecting  three nodes) are defined as basic graphs 
as they are the smallest meaningful graphs that can model composition patterns. They are grouped as: A) \textit{transitive}: 
directed primitive relations are connected in a uniform direction;
B) \textit{common source}: the starting node of two directed primitive relations are shared; C) \textit{common target}: the end nodes of two primitive directed relations are shared. Fig. \ref{fig:subgraph} illustrates these connection schemas. 

\begin{table*}
    \centering
    \resizebox{.77\columnwidth}{!}{
    \begin{tabular}{ll|ccccc|cc|c} \toprule
         &Dimension &\multicolumn{2}{c}{ROUGE-2/L} &\multicolumn{2}{c}{BLEU-3/4} &METEOR &CIDEr &SPICE &Coverage \\ \midrule
         \multirow{3}{*}{\rotatebox{90}{level}}&easy (typeA) &5.99 &23.52 &10.50 &5.80 &13.96 &6.10 &20.10 & 66.63 \\
         &medium (typeB) &5.69 &23.39 &9.90 &5.40 &14.01 &6.12 &20.60 &66.43 \\
         &hard (typeC) &5.67 &22.98 &10.00 &5.50 &14.13 &6.01 &21.10  &66.58 \\ \midrule
         \multirow{2}{*}{\rotatebox{90}{order}}&easy-to-hard &\textbf{7.11} & \textbf{27.77} & \textbf{12.00} &\textbf{7.10} & \textbf{16.76} & \textbf{7.65 }& \textbf{27.20} & \textbf{79.98} \\
         &hard-to-easy&7.04 &26.60 &11.50 &6.80 & 15.61 & 7.25 &24.40 & 73.06 \\ \bottomrule
    \end{tabular}}
    \caption{Performance results when 
    controlling the demonstration of graph structures in ICL along two dimensions: \textit{level of difficulty} (level) and \textit{ordering according to difficulty} (order). The pairwise t-test at 5\% significance level. 
    }
    \label{tab:rst_order}
\end{table*}

We categorize the graph structures of test samples
along the classes A-C.
Table \ref{tab:rst_type} shows the performance gap between in-distribution and compositional generalization for the 3 classes.
\footnote{We compare results of each two data structures, the pairwise t-test at 5\% significance level.} We observe that \textbf{different structures present varying levels of difficulty for compositional generalization}, with A) \textit{transitivity} < B) \textit{common source} < C) \textit{common target}. 
We further computed the occurrence frequency of subgraphs of the different composition types, with detailed statistics provided in Appendix \ref{app:stat_graph_type}. 
It shows that Type A occurs more frequently than Types B and C, while the complexity of the three structures in terms of number of nodes and edges is comparable. 
We speculate that the difference in difficulty is likely because, compared to a common source or target structure, transitive structures are more common and straightforward to construct into a natural sentence.

\begin{figure}
    \centering
    \includegraphics[width=\linewidth]{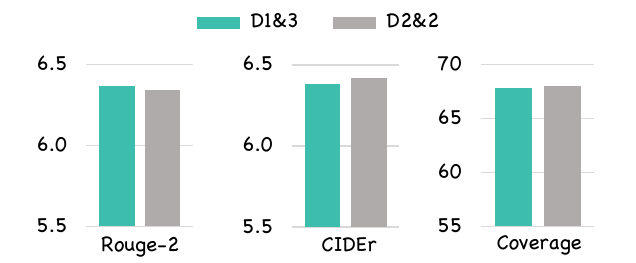}
    \caption{Two ways of providing the
    required primitive relations in demonstrations for in-context learning.}
    \label{fig:demo-type}
\end{figure}

\paragraph{Structuring Demonstrations}
Although we guarantee that primitive relations required to solve unseen composition samples are fully covered by in-context demonstrations, the distribution of these primitive relations in ICL demonstrations is not fully constrained. For example, for a test sample with graph label \textit{Causes \& IsA \& HasProperty \& UsedFor},
the 
demonstrations could provide primitive relations in the following two ways: i) D$_{1\&3}$. The compositional relations could be 
separated into one relation and the three remaining ones,  such as samples with \textit{Causes} and \textit{HasProperty \& IsA \& UsedFor}; 
or ii) D$_{2\&2}$. The compositional relations could be separated into two groups with  
two primitive relations each, such as \textit{Causes \& HasProperty} and \textit{IsA \& UsedFor}.
Figure \ref{fig:demo-type} illustrates the comparison between these two alternative
ways of providing
the required primitive relations. \footnote{Given space limits, we select a representative metric from each metric category.
Other choices show a similar trend.} We observe that the alternative options in structuring ICL demonstrations in terms of packaging primitives show
minor differences. This indicates that alternative options for presenting \textit{the same set of primitives} for a given sample do
not affect the performance of compositional reasoning significantly. 

\subsection{Primitive Analysis}
\label{ana:prim}

Next, we examine the correlation between primitive relation types and the models' performances in the two data configurations, aiming to determine whether specific primitive relations have a specific impact (i.e., greater or lower difficulty) on a model's 
compositional generalization abilities. Specifically, we count the occurrence frequency of all primitive relation types 
in the reasoning graphs, and rank these relations from most to least frequent. The sorted rank is denoted as \textit{R$_{freq}$}. We also examine the difficulty of primitive relation types based on the two data configurations (id vs.\ ood) from easiest to hardest, and record
the gap in performance ($\Delta$) between the two configurations from smallest (best, highest rank) to largest (worst, lowest rank). Each rank for a given relation under a given data configuration is compared with its frequency rank \textit{R$_{freq}$}. Table \ref{tab:rst_prim_part} shows Rank Biased Overlap (RBO) \citep{webber2010similarity} results for three metrics and their average value. 

\begin{table}
    \centering
    \resizebox{.9\columnwidth}{!}{
    \begin{tabular}{c|ccc|c} \toprule
         Dis &RBO$_{R2}$ &RBO$_{CIDEr}$ &RBO$_{Cov}$ & Avg\\ \midrule
         id &0.5445	&0.5736	&0.5141	&0.5441 \\
         ood &0.4927	&0.5473	&0.5119	&0.5173 \\ 
         $\Delta$ &0.4603 &0.3997 &0.4997 &0.4532 \\ \bottomrule
    \end{tabular}}
    \caption{Rank Biased Overlap (RBO) compares the rank of relation types according to 
    their frequency and their associated performance
    results across three data configurations (Dis). Avg indicates the average results.}
    \label{tab:rst_prim_part}
\end{table}

We find that the average correlations of the three configurations are around 50\%, indicating a moderate correlation between performance and the frequency of primitive relation types. Notably, 
the \textit{id} and \textit{ood} configurations show correlations of 0.5441 and 0.5173, respectively, which are higher than the 0.4532 correlation observed for gap performance. This suggests that primitive relation types have a greater impact on absolute generation quality than on a model's 
compositional generalization abilities.

\section{Difficulty-based Demonstrations}
We conclude from the analyses in Sec.\ \ref{analyses} (see especially Table \ref{tab:rst_type}) that reasoning graph structures are a significant factor affecting
compositional generalization. Hence, we aim to investigate whether, and to what extent the compositional generalization ability of evaluated LLMs could be enhanced by controlling the demonstration of graph structures 
in in-context learning. We group 
the samples by the ranked difficulty of their graph structures into three groups: 
\textit{hard}: type C, \textit{medium}: type B, and \textit{easy}: type A (see 
Fig.\ \ref{fig:subgraph}). We then 
select and arrange demonstration candidates along  
two dimensions: i) \textbf{level of difficulty}, by selecting demonstrations from a specific graph structure type (A/B/C) and
ii) \textbf{ordering according to difficulty}, 
by arranging
demonstration types according to a given level of difficulty and following a specific order, from \textit{easy-to-hard} (A$\rightarrow$B$\rightarrow$C) or \textit{hard-to-easy} (C$\rightarrow$B$\rightarrow$A).

Table \ref{tab:rst_order} shows results for both dimensions. 
We find that the evaluated model benefits more when demonstrations are \textit{ordered by difficulty}.
That is, combining graph structures of different difficulty levels considerably enhances the model's ability to perform compositional reasoning -- compared to relying on a single structure type. This finding aligns with results in \citet{levy-etal-2023-diverse} who experimented on tree structures. Furthermore, we observe ordering demonstrations in an \textit{easy-to-hard} manner achieves superior compositionality performance compared to the reverse demonstration order. 
This result parallels the findings of \citet{fu2024exploring}, who show that ordering compositional textual NLI problems in an easy-to-hard manner improves model performance in continual learning.

\section{Conclusion}
We propose a Compositional Generalization challenge for Graph-based Commonsense Reasoning that extends CommonGen to a compositional \textit{generative commonsense reasoning} task from \textit{graph-structured inputs}. Extensive experiments on seven LLMs using In-Context Learning indicate that they struggle with compositional generalization settings. We investigate potential causes of the limitations, and find that the topology of the graph structures is a significant factor. We show that arranging the order of demonstrations in an easy-to-hard schema enhances the compositional generalization ability.

\section{Limitations}
We use ConceptNet to enrich each CommonGn sample (a concept set and reference sentence) with a commonsense reasoning graph, which constrains potential relations between concepts to 13 common commonsense relation types. This limitation restricts the variety of composition types compared to real-world applications. However, even on this restricted set of basic relations we were able to establish weaknesses of current LLMs. Additionally, the construction of the data is unsupervised and relies on the quality of ConceptNet and the SBERT model. However, the proposed construction method for CGGC is flexible and can be extended with other high-quality and high-coverage commonsense resources in the future. 

For in-context learning, we use a fixed prompt as described in Appendix \ref{app:prompts}, chosen based on recommended best practices and preliminary experiments. We leave the exploration of other prompt constructions, such as incorporating explanations within the prompts, for future work.

\section{Acknowledgments}
We are grateful to anonymous reviewers for their valuable comments that have helped to improve this paper. We also thank annotators for their valuable work on human evaluations. This work has been supported through a scholarship provided by the Heidelberg Institute for Theoretical Studies gGmbH.

\bibliography{anthology,custom}
\bibliographystyle{acl_natbib}

\clearpage

\appendix
\section{Data}
\subsection{ConceptNet Relations}
\label{app:cn_rels}
We merge some relations with similar meanings, e.g., \{/r/InstanceOf, /r/MannerOf\} $\rightarrow$ /r/IsA. Table \ref{tab:cn_rels} shows all instantiations of merged relations. We also ignore some infrequent relations, e.g., /r/LocatedNear, /r/SymbolOf. Finally selected 13 relations are listed in the bottom of Table \ref{tab:cn_rels}.

\begin{table}[h]
\centering
\resizebox{\columnwidth}{!}{
\begin{tabular}{@{}lp{3.5cm}p{3.5cm}@{}} \toprule
\multirow{8}{*}{\rotatebox{90}{merge\_rel}} & source &target \\ \cmidrule(r){2-3}
 & /r/HasFirstSubevent, /r/HasLastSubevent& /r/HasSubevent \\ \cmidrule(r){2-3}
 &/r/InstanceOf, /r/MannerOf &/r/IsA \\ \cmidrule(r){2-3}
 &/r/PartOf &/r/HasA \\ \midrule\midrule
\multirow{4}{*}{\rotatebox{90}{final\_rel}} & \multicolumn{2}{p{7.5cm}}{/r/IsA, /r/UsedFor, /r/AtLocation, /r/HasSubevent, /r/HasPrerequisite, /r/CapableOf, /r/CausesDesire, /r/Causes, /r/MotivatedByGoal, /r/HasProperty, /r/ReceivesAction, /r/HasA,  /r/Desires}
 \\ \bottomrule
\end{tabular}}
\caption{Instantiations of merged relations and finally selected relations from ConceptNet.}
\label{tab:cn_rels}
\end{table}

\subsection{Data Statistics of Graph Types}
\label{app:stat_graph_type}
Considering the complexity of compositions with multiple primitives, we constrain the experiment to two primitives. Experimental data is sampled from graphs of original concepts set size 3.
Selected test samples are categorized given their graph structures, as illustrated in Fig.\ \ref{fig:subgraph}. Table \ref{tab:type_statistics} presents the data statistics for the three different types of graph structures. We observe that type A occurs more frequently than 
the other graph types.
However, the graph complexity in terms of node and relation counts is comparable, with Type A showing marginally lower counts.

\begin{table}[h]
    \centering
    \resizebox{\columnwidth}{!}{
    \begin{tabular}{c|ccccc} \toprule
         &\multicolumn{2}{c}{general} &\multicolumn{3}{c}{graph} \\ \cmidrule(r){2-3} \cmidrule(r){4-6}
         & \#num & \#sent & \#nodes & \#edges &\#rel\\ \midrule
        Type A &3517 &6250 &8.49 &7.17 &4.41   \\
        Type B &1037 &1697 &9.22 &7.91 &4.72   \\
        Type C &1185 &1964 &9.12 &7.86 &4.73   \\ \bottomrule
    \end{tabular}}
    \caption{Data statistics of different graph types of test samples in CGGC. We include the general information of concept sets and extended reasoning graph details.}
    \label{tab:type_statistics}
\end{table}

\subsection{Data Statistics of Target Sentences}
\label{app:stat_target_sent}
We segmented the target sentences and counted the involved words. The column \#len$_s$ in Table \ref{tab:stat_sent} presents the results, showing that each sentence contains an average of 11 words. We further explored the new words required to generate the target sentence. Specifically, we removed stop words in each sentence. New words are defined as follows: i) \textit{w/o graph},  only given concepts are counted as given words; ii) \textit{w/ graph}, 
concepts contained in the graph are also counted as given words. The column \#nw$_{wg}$ in Table \ref{tab:stat_sent} indicates that a generated sentence requires roughly 3 new meaningful words. We also analyze the part-of-speech tagging of these novel words. C$_{nw}$@5 shows the top 5 categories of missing words. It shows that the categories of missing words mainly include verbs, nouns, adjectives, and prepositions. 

\subsection{Data Statistics of Graph Labels}
\label{app:stat_graph_label}
\begin{figure}[h]
    \centering
    \includegraphics[width=\linewidth]{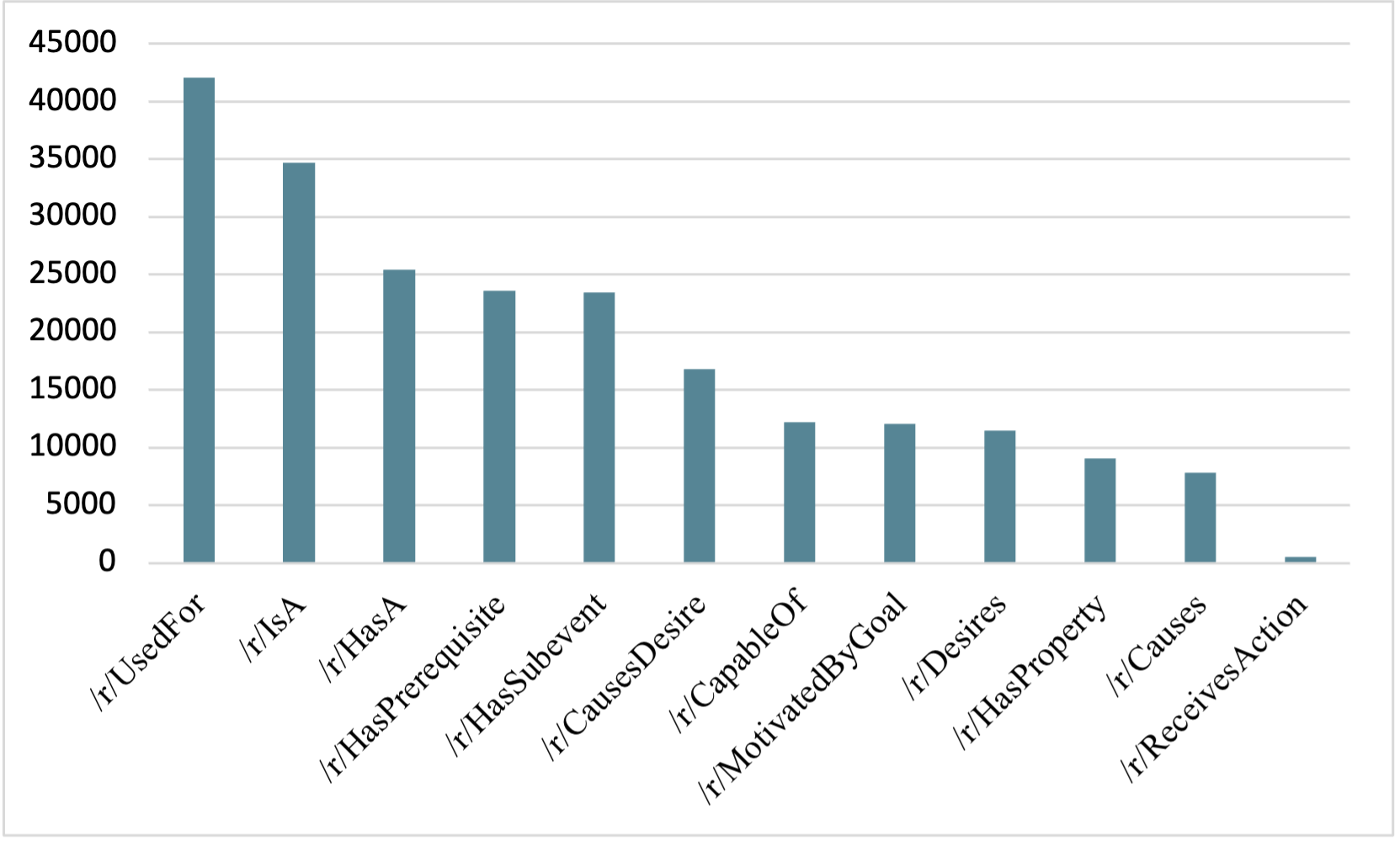}
    \caption{Distribution of primitive relations. The y-axis indicates number of samples.}
    \label{fig:relation_dis}
\end{figure}

\begin{figure*}
    \centering
    \includegraphics[width=1\linewidth]{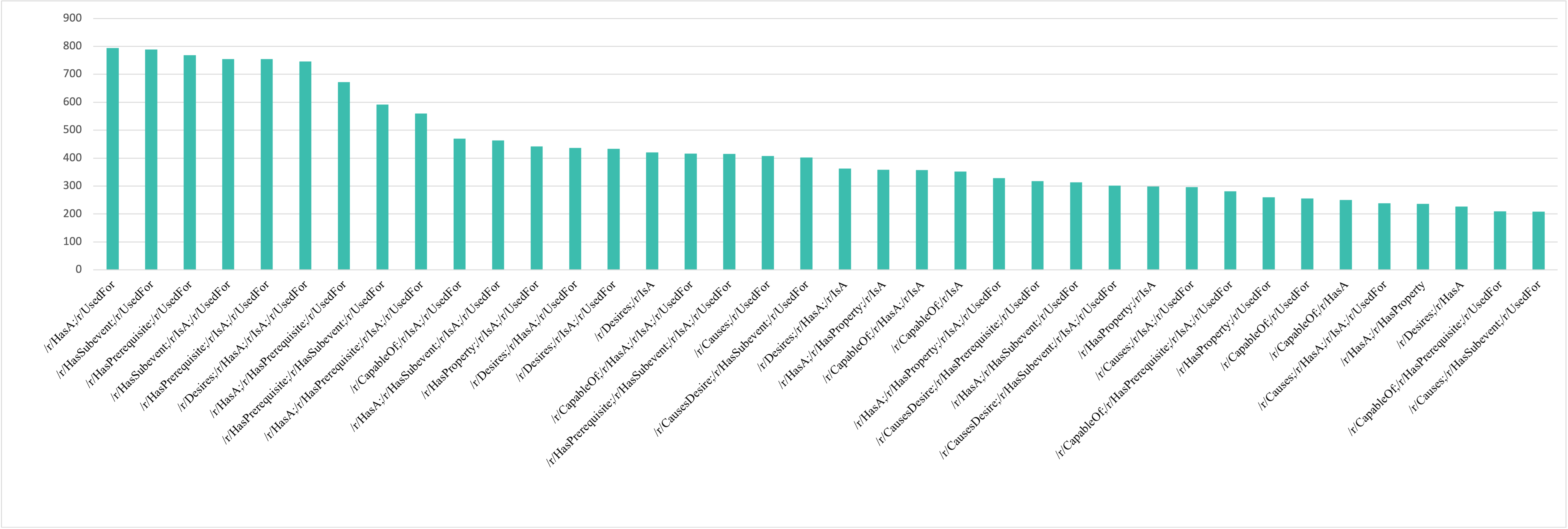}
    \caption{Distribution of compositional \textit{graph labels}. The y-axis indicates number of samples.}
    \label{fig:dis_graph_labels}
\end{figure*}

As mentioned in Section \ref{sec:graph_label}, we ultimately selected 468 compositional graph labels for our experiments. Figure \ref{fig:dis_graph_labels} illustrates the sample distribution for the top 40 graph labels. We further count various relation types (primitives), where Figure \ref{fig:relation_dis} show the sample distribution for used relations.

\begin{table*}[h]
    \centering
    \resizebox{.9\columnwidth}{!}{
    \begin{tabular}{c|ccccc} \toprule
        &\multirow{3}{*}{\#len$_s$} &\multicolumn{2}{c}{w graph}&\multicolumn{2}{c}{w/o graph}\\ \cmidrule(r){3-4}\cmidrule(l){5-6}
        &  &\#nw$_{wg}$& C$_{nw}$@5 & \#nw$_{wog}$ & C$_{nw}$@5 \\ \midrule
        all &10.68 &2.82&VBG,CD,NN,JJ,IN &2.87&VBG,CD,NN,JJ,IN \\ \midrule
        -size=3  &10.16 &2.81&VBG,CD,NN,JJ,IN &2.86&VBG,CD,NN,JJ,IN \\
        -size=4  &11.75  &2.83&RB,JJ,IN,NN,NNS &2.88&RB,JJ,IN,NN,NNS \\
        -size=5  &13.89 &2.98&NN,JJ,NNS,VBG,VBD &3.05&NN,JJ,NNS,VBG,VBD \\ \bottomrule
        \multicolumn{6}{l}{\makecell[l]{\textbf{VBG}: Gerund or Present Participle; \textbf{CD}: Cardinal Number; \textbf{NN}: Noun, Single; \textbf{JJ}: Adjective; \\\textbf{IN}: Preposition; \textbf{NNS}: Noun, Plural; \textbf{RB}: Adverb; \textbf{VBD}: Verb, Past Tense}} \\ \bottomrule
    \end{tabular}}
    \caption{Data statistics for the target sentences of the CGGC Benchmark. 
   \textit{\#nw} indicates the number of novel words (removing stop words and punctuation) except the given concepts contained in the target sentence.
    The subscript $_{wg}$ and $_{wog}$ denotes whether concepts contained in the reasoning graph are counted in known concepts. C$_{nw}$@5 denotes the top 5 categories of missing words.}
    \label{tab:stat_sent}
\end{table*}

\subsection{Data Splits for \textit{Verification} \& \textit{Composition}}
\label{app:data_split}
As mentioned in Section \ref{verification}, we need to construct a verification model \footnote{For details of the verification model see Appendix \ref{app:verification}{}.} to verify if the generated sentence uses the reasoning relations provided in the graph as we expected. Hence, we split the dataset into two groups: i) \textit{verification}, for evaluating whether the generated sentences do in fact express the target relations. Train and val data are used for verification model training and validation. Results are shown in Appendix \ref{app:verification};
and ii) \textit{composition}, for graph-based compositional generalization tests. Here the train and test data are used for constructing demonstrations and compositional test. Results are shown in Section \ref{main_rst}. Table \ref{app:data_statis} shows the data statistics for the above two groups.

\begin{table}[h]
 \resizebox{0,8\columnwidth}{!}{
\begin{tabular}{lccccc} \toprule
& \multicolumn{2}{l}{Verification}& & \multicolumn{2}{l}{Composition} \\ \cmidrule{2-3} \cmidrule{5-6}
 &train&    val&  & train& test \\ \midrule
\#set &6375&    1125&   &11591& 7728      \\ \bottomrule  
\end{tabular}}
\caption{Data statistics for verification and composition. The sample counts are based on the number of concept sets.}
\label{app:data_statis}
\end{table}

\begin{table*}
    \centering
    \resizebox{0.8\columnwidth}{!}{
    \begin{tabular}{l|ccccccc} \toprule
         Dis & Llama2& Llama2-chat & Mistral & Mistral-chat & Falcon & Falcon-chat & GPT-J \\ \midrule
         id&44.64& 44.53& 45.11& 44.80& 44.47& 44.27& 45.33 \\
         ood&44.42& 44.69& 45.25& 45.28& 44.44& 44.18& 45.43 \\
         \bottomrule
    \end{tabular}}
    \caption{The filtering rate (\%) by the verification model for seven evaluated models under two test configurations in-distribution (id) and compositional generalization (ood).}
    \label{tab:rst_verification_rate}
\end{table*}

\section{Human Evaluation of Reasoning Graphs} 
\label{app:graph_eval}
As our reasoning graphs are automatically constructed, we perform a human evaluation to assess whether these graphs are related to the target sentence, following \cite{josifoski-etal-2023-exploiting}. We randomly selected 50 test samples for this evaluation. We hired two annotators who majored in computational linguistics for annotation. For each sample, annotators were presented with the target sentence and 
all triples extracted from the corresponding graph. Each triple is comprised by two concepts and their relation. For each triple, 
annotators were instructed to determine if the relation between the two given concepts could be inferred from the target sentence. We annotate all triples in one sample, and each sample was rated on a scale of 0 (not related), and 1 (related). The percentage agreement between annotators is 96\%, with a Cohen's kappa of 81.13\%. 
The results demonstrate a 91\% accuracy in the extracted subgraph's relevance to the target sentence, indicating the high quality of the constructed graphs.

\begin{table*}
    \centering
    \resizebox{\columnwidth}{!}{
    \begin{tabular}{l|l} \toprule
         Quality & \makecell[l]{\underline{Is the sentence well-formed well-formed?} \\ \textit{Yes}: The sentence is well-formed and fluent. \\ \textit{Somewhat}: The sentence is understandable but a bit awkward. \\ \textit{No}: The sentence is neither well-formed or fluent.}\\ \midrule 
         Plausibility &\makecell[l]{\underline{Does the sentence describe a plausible scenario?} \\ \textit{Yes}: The sentence describes a realistic or plausible scenario. \\ \textit{Somewhat}: The sentence describes an acceptable scenario but a bit awkward. \\ \textit{No}: The sentence describes a nonsensical scenario. } \\ \midrule
         Concepts &\makecell[l]{\underline{Does the sentence include the given concepts meaningfully?} \\ \textit{Yes}: The sentence meaningfully includes all of the concepts. \\ \textit{Somewhat}: The sentence meaningfully includes some, but not all of the concepts. \\ Or, the sentence includes all concepts but some of them are not meaningful or properly incorporated. \\ \textit{No}: The sentence does not include concepts in a meaningful way. } \\ \midrule
         Overall &\makecell[l]{\underline{Considering your answers to 1), 2) and 3), does the sentence meaningfully combine all of the concepts} \\ \underline{into a well formed and plausible scenario?} \\ \textit{Yes}: The sentence is reasonably understandable, and meaningfully combines all the concepts into a plausible scenario. \\ \textit{Somewhat}: The sentence looks okay in terms of above questions. \\ \textit{No}: The sentence is not well-formed/understandable, or fails to properly combine all the concepts into a plausible scenario. } \\
         \bottomrule
    \end{tabular}}
    \caption{Human evaluation guidelines for evaluating the generated sentences.}
    \label{tab:guideline}
\end{table*}

\section{Experiment Details}
\label{app:prompts}
\paragraph{Prompts} To guide the given task, we add an instruction at the beginning of all inputs. We provide a prompt example as follows:
\begin{center}
    \small
    \begin{minipage}{0.45\textwidth}
        \textit{Please generate a natural sentence with the provided concepts and their commonsense reasoning graphs.} \par
        \textit{concepts: oven, cheese, pizza} \par
        \textit{commonsense reasoning graph: <H> pizza <R> HasA <T> cheese, <H> pizza <R> AtLocation <T> oven} \par
        \textit{sentence:} \par
    \end{minipage}
\end{center}
Following the requirements of LLMs, we also add some special tokens to the prompts: i) for \textit{Llama2} models, we add `[INST] <<SYS>>' (detailed templates can be found in the official website \footnote{\url{https://llama.meta.com/docs/model-cards-and-prompt-formats/meta-llama-2/}}); ii) for \textit{Falcon} and \textit{Mistral} models, we add `User:' and `Assistant:'.

\section{Verification Model}
\label{app:verification}
We aim to construct a \textit{verification model} to ensure that the generated sentence accurately employs the required reasoning relations. This model is a multi-label classifier based on a LLM. Specifically, we use Llama2 along with a feed-forward classification layer. The model classifies the relation type based on two concepts and a target sentence. For instance, given the concepts \{\textit{oven, pizza}\} and the target sentence \textit{He puts a cheese pizza in an oven}, we expect the model to predict \textit{/r/AtLocation}. To test the model, we compare the predicted relation with the relations specified in the reasoning graph. If the predicted relation matches the provided relation, we consider this relation to have been genuinely used, aligning with our expectations for compositional evaluation.

Table \ref{app:data_statis} presents the data statistics for the verification model. We use 6735 samples for training, achieving 90.21\% accuracy on a validation set of 1125 samples. To ensure data quality for verification, we sampled data for human evaluation. The results in Appendix \ref{app:graph_eval} confirm that the data is suitable for verification. 

We use this verification model to filter the results of the compositional generation test. When doing so, we focus exclusively on examples where all composed relations are accurately applied, i.e., samples that achieved 100\% verification accuracy. Table \ref{tab:rst_verification_rate} shows the filtering ratio (\%) for each model in the two evaluation configurations: in-distribution and compositional generalization.

\begin{figure*}
    \centering
    \includegraphics[width=1\linewidth]{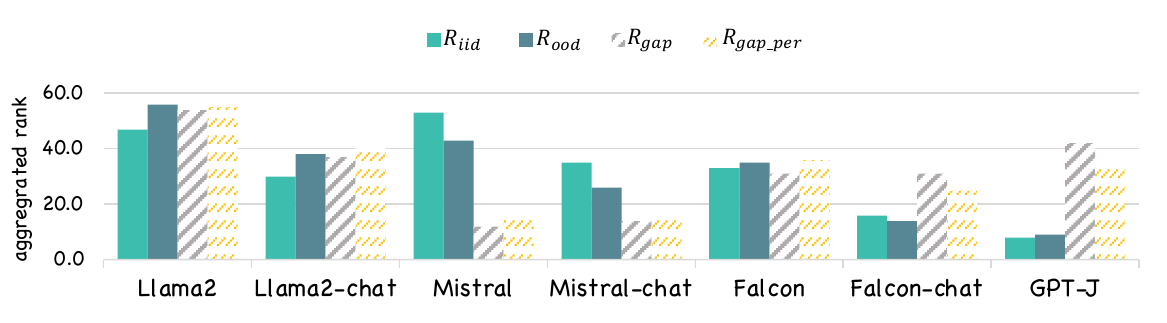}
    \caption{Aggregated results based on eight evaluation metrics for seven evaluated LLMs. Under each evaluation metric, LLMs' results are ranked in an ascending order. The results for \textit{id} and \textit{ood} represent absolute performance, i.e., higher rank is better. Conversely, the \textit{gap} and \textit{gap percentage} show gap performance, i.e., lower rank is better. We then score LLMs based on their ranks. Scores for \textit{id} and \textit{ood} are assigned from 1 (lowest=worst) to 8 (highest=best), while for \textit{gap} and \textit{gap percentage} are assigned from 8 (highest=best) to 1 (lowest=worst). For each model of a specific configuration, we summarize the rank scores across the eight metrics (minimum value: 8*1=8, maximum value: 8*7=56).
    }
    \label{app:fig_aggregrated_rst}
\end{figure*}

\section{Human Evaluation Guidelines}
\label{app:huamn_eval_guidline}
Following \citep{lu-etal-2022-neurologic, meng2022controllable,zhang2023tractable}, we conduct a human evaluation of the generated sentences $y$ across four dimensions. The guidelines of these four aspects for annotators are provided in the Table \ref{tab:guideline}.

\section{Aggregated Results}

\label{app:aggregated_rst}
Figure \ref{app:fig_aggregrated_rst} aggregates the main results shown in Table \ref{tab:main_rst} for better illustration. Under each evaluation metric, we ranked and scored the LLMs' results for a specific configuration in a specific order. Under each evaluation metric, LLMs’ results are ranked in an ascending order. The results for id and ood represent absolute performances, that is, higher rank is better. Conversely, the gap and gap percentage (the percentage drop of the gap value in relation to id results, i.e., smaller is better) represent relative gap performance, that is, lower rank is better. We then score seven LLMs based on their ranks. For absolute performance under id and ood configurations, scores are assigned from 1 to 7. Conversely, scores for gap performance are assigned from 7 to 1. For example, under the Rouge-2 evaluation for the \textit{id} data configuration, the seven LLMs will be ranked and scored as: (7) Mistral > (6) Llama2 > (5) Falcon > (4) Mistral-chat > (3) Llama2-chat > (2) Falcon-chat > (1) GPT-J. For each model of a specific configuration, we summarize the rank scores across the eight metrics. The minimum score is eight metrics with score one, i.e., 8*1 =8, and the maximum value is eight metrics with score seven, i.e., 8*7=56. Aggregated main results are illustrated in Fig. \ref{app:fig_aggregrated_rst}.


The aggregated ranks under in-distribution (R$_{id}$) and compositional generalization (R$_{ood}$) represent the absolute generation ability, while the remaining two ranks based on the gap performance (R$_{gap}$) and gap percentage(R$_{gap\_per}$) indicate the models' compositional generalization capability.
Given the aggregated scores shown in Fig. \ref{app:fig_aggregrated_rst}, we find Mistral achieves the best performance under in-distribution and Llama2 obtains the best results under compositional generalization and smallest gaps. Further, we observe that \textbf{various model groups show variance in absolute performance and compositional generalization}. Llama2-based models (Llama2 and Llama2-chat) and Falcon show consistent ability in generation and compositional generalization, while Llama2 represents the top ability. In contrast, the remaining LLMs show a difference between these two abilities. Mistral-based models achieve relatively high absolute performance but low compositional generalization capability, whereas GPT-J shows the opposite trend.

\end{document}